\documentclass[lettersize,journal]{IEEEtran}
\usepackage{amsmath,amsfonts}
\usepackage{algorithmic}
\usepackage{algorithm}
\usepackage{array}
\usepackage[caption=false,font=normalsize,labelfont=sf,textfont=sf]{subfig}
\usepackage{textcomp}
\usepackage{stfloats}
\usepackage{url}
\usepackage{verbatim}
\usepackage{graphicx}
\usepackage{cite}
\usepackage{booktabs}
\usepackage{multirow} 
\usepackage{capt-of}

\usepackage[table]{xcolor}
\definecolor{HOColor}{HTML}{00A2FF}
\definecolor{HEColor}{HTML}{60D935}

\newcommand{\hohe}{\textsc{Ho-He}\xspace}
\newcommand{\ho}{\textsc{Ho}\xspace}
\newcommand{\he}{\textsc{He}\xspace}

\newcommand{\ie}{\textit{i.e.,}\@\xspace}
\newcommand{\eg}{\textit{e.g.,}\@\xspace}

\newtheorem{proposition}{Proposition}

\begin{document}

\title{Post-Generation Curation of Synthetic Images via
Homogeneous-Heterogeneous Splitting}

\author{Disheng Liu, Tuo Liang, Chaoda Song, and Yu Yin%
\thanks{Disheng Liu, Tuo Liang, Chaoda Song, and Yu Yin are with the
Department of Computer and Data Sciences, Case Western Reserve University,
Cleveland, OH, USA (e-mail: \{dxl952, txl859, cxs965, yxy1421\}@case.edu).}%
\thanks{Corresponding author: Yu Yin.}}




\maketitle

\begin{abstract}
Recent generative models can produce high-quality synthetic images, offering scalable training data for data-hungry models. Existing approaches to exploiting this potential typically involve 1) training or fine-tuning generators, or 2) using lightweight post-hoc adaptation like prompt engineering or inference-time guidance, making them generator-specific and expertise-intensive. We study a complementary question: \emph{given a fixed pool of generated images, can downstream utility be improved purely by selecting an informative subset?} The answer is yes. We show that effective selection must counter a structural bias of modern generators: they tend to over-produce canonical modes of each class while under-representing intra-class variation. Building on this insight, we split real data into a \emph{canonical Homogeneous} (\ho) \emph{subset} and a \emph{non-redundant Heterogeneous} (\he) \emph{subset}, then score synthetic images by a fidelity-diversity criterion that rewards semantic alignment while penalizing canonical redundancy. The method is generator-agnostic and does not require retraining. Across multiple benchmarks, it consistently outperforms state-of-the-art data selection baselines and matches the real-data performance with up to 40\% fewer synthetic samples. The same criterion remains effective when applied on top of stronger task-tuned generators, with gains on both classification and segmentation tasks. Thus, such post-generation selection acts as a complementary mechanism for better generators, improving the utility of synthetic data.
\end{abstract}


\section{Introduction}
\IEEEPARstart{H}{igh-quality} synthetic images offer a promising solution for data scarcity in data-intensive AI. Recent studies demonstrate that generative models (GM) can effectively replicate datasets (\eg CIFAR-10, ImageNet) and improve downstream models trained on such synthetic data in generalization, transferability, and in-domain accuracy~\cite{zheng2023understandinggenerativedataaugmentation, sariyildiz2023faketillmakeit, azizi2023syntheticdatadiffusionmodels, he2023syntheticdatagenerativemodels, li2024syntheticimageusefultransfer, fan2023scalinglawssyntheticimages}.

However, practical challenges remain: 1) synthetic datasets may contain low-fidelity or mislabeled samples, introducing harmful noise; 2) Inadvertently over-representing dominant patterns could transfer inherent biases to downstream models; 3) While scaling the data alleviates these issues, it incurs higher computational overhead and longer training times. 

To harness synthetic data, researchers are prioritizing improved generation techniques that yield more faithful and diverse data. These approaches can be broadly categorized into: 1) training- or fine-tuning-based methods~\cite{azizi2023syntheticdatadiffusionmodels, kim2024datadream}, which adapt generators to better match target distributions; 2) lightweight strategies like guidance~\cite{yuan2024realfakeeffectivetrainingdata, kim2025test, liang2025diffusion, dall2025increasing, kirchhof2024shielded} or prompting~\cite{Shipard_2023, rahat2025data, dunlap2023diversify}. While effective, these approaches face distinct trade-offs: training-based methods incur substantial computational overhead due to the optimization of large-scale models, whereas lightweight strategies often rely on model-specific configurations, significant domain expertise, or extensive empirical tuning. Alternatively, data selection methods~\cite{gadre2023datacomp,abbas2023semdedup} enhance dataset quality through post-hoc curation instead of generation-time steering. This data-centric focus complements task-guided generation by further improving synthetic quality.

\begin{figure}
  \centering
  \includegraphics[width=\linewidth]{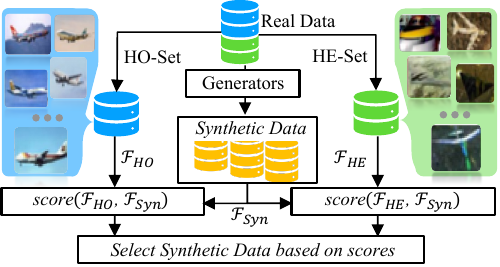}
  \caption{Data selection by considering both fidelity and diversity. In the pipeline, real data (CIFAR-10 \textit{airplane}) is divided into the \ho- and the \he-Set. Synthetic data is scored and selected by referring to such two partitions.}
  \label{fig:whole_pipeline}
\end{figure}

Aligning curated data with the targeted distribution is a promising post-processing principle. Image–Label Alignment methods~\cite{he2023syntheticdatagenerativemodels,chen2023effective, zhang2021datasetgan} assume that high-quality data align strictly with the generative labels, employing pre-trained discriminators to filter out noisy samples; Image–Image Alignment~\cite{NEURIPS2019_0234c510, lin2023explore} prioritizes synthetic samples that closely resemble real images. Both strategies emphasize fidelity, but largely neglect diversity, reducing the utility of synthetic data~\cite{fan2023scalinglawssyntheticimages} as repeated patterns and missing novel information.

In this work, we propose a \textit{fidelity–diversity} balanced post-generation curation framework to enhance the utility of synthetic imagery for downstream tasks.
During the curation, we first split the real data into two subsets: a \textbf{Homogeneous} set characterized by high internal similarity, and a \textbf{Heterogeneous} set enriched with variation. Based on this partitioning, we design a scoring mechanism to identify and select desired synthetic instances, balancing high semantic quality with representational diversity, shown as Fig.~\ref{fig:whole_pipeline}. Concretely, we compute two complementary metrics: 1) \textit{fidelity scores}, measuring semantic similarity to real samples; 2) \textit{diversity scores}, quantifying the deviation from repetitive patterns. Combining such metrics, we curated the final desired synthetic pool.

To evaluate our curation strategy, we conduct extensive experiments across diverse datasets (e.g., CIFAR-10 and ImageNet) and model architectures (e.g., ResNet and ViT). We construct synthetic data pools using generators trained on the target distributions and curate these pools for downstream model training. We further apply our curation strategy to customized generators designed to enhance the downstream utility of synthetic data, demonstrating its complementarity, as a post-generation curation, to task-guided synthesis in both classification and segmentation settings. Finally, we evaluate the trained models under out-of-distribution settings, demonstrating the robustness of models trained on curated synthetic data.
In summary, our contributions are as follows:

\begin{enumerate}
\item A nearest-neighbor-cover partition (\hohe) of real data with a minimality guarantee (Prop.~\ref{prop:nn-cover-main}), giving an explicit notion of canonical vs.\ non-redundant samples.
\item A principled, post-generation selection strategy that jointly quantifies and balances fidelity and diversity. The framework is generator-agnostic, requiring only a synthetic data pool, and avoids costly retraining or fine-tuning of generators.

\item Extensive experiments validate the approach across classification and segmentation benchmarks with multiple generators and backbones. Our method consistently improves in-domain accuracy and OOD robustness over baselines, plus an additional plug-in evaluation on top of generator-side intervention methods.

\end{enumerate}

\section{Related Work}
\label{sec:EWs}

\subsection{Curating Synthetic Data for downstream utility}
Synthetic data has emerged as a promising solution to data scarcity in the AI era~\cite{tang2025aerogen, qiu2025noise, wang2024domain}. Recent studies show that models trained on synthetic data can learn robust visual representations~\cite{he2023syntheticdatagenerativemodels, zhou2023trainingairimproveimage, 
tian2023stablerepsyntheticimagestexttoimage, hammoud2024synthclipreadyfullysynthetic, tian2023learningvisionmodelsrivals}. Moreover, augmenting real datasets with synthetic samples has been shown to further improve model performance~\cite{sariyildiz2023faketillmakeit, zeng2025contrastivelearningsyntheticpositives, qraitem2024fakerealpretrainingbalanced}. However, the distribution gap between real and synthetic data~\cite{fan2023scalinglawssyntheticimages, Hataya_2023} highlights the need for careful tailoring~\cite{li2024syntheticimageusefultransfer, wang2024domain, wang2024generateddatahelpcontrastive, guo2026utilgen} to fully exploit the potential of synthetic data.

Intervening before or during generation is an effective way to improve the utility of synthetic data. Fine-tuning-based methods adapt generators to target distributions or downstream tasks~\cite{azizi2023syntheticdatadiffusionmodels, kim2024datadream}, while prompt-, guidance-, and sampling-based methods steer generated samples toward desired classes, attributes, or harder variations~\cite{yuan2024realfakeeffectivetrainingdata, kim2025test, liang2025diffusion, rahat2025data}. Although such approaches can improve the utility of synthetic data, they often require generator access, task-specific tuning, prompt expertise, or additional generation cost. In contrast, post-generation curation offers a practical and scalable alternative to generation-time intervention for improving data utility.

\subsection{Post-Generation Synthetic Data Curations}
Unlike real data pruning \cite{abbas2023semdedup, abbas2024effective, chandhok2025learning}, synthetic data involves generator bias, distribution drift, and mode collapse, thus, requiring careful curation for downstream usage.

\noindent \textbf{Fidelity-Guided Curation.} High-fidelity generations ensure semantic correctness, making fidelity-based selection effective. Such curation strategies can be broadly categorized into two main approaches: \textbf{1) Image-Label Alignment:}
pretrained or task-specific models are used to filter low-quality samples by discarding those misclassified within top-$k$ predictions~\cite{Bhattarai_2020, Zhang_2021, chen2023data, Vu_2021}. Concretely, CLIP~\cite{pmlr-v139-radford21a} is used to assess image-label semantic alignment~\cite{he2023syntheticdatagenerativemodels, lin2023explore, tang2025trainingfreesyntheticdataselection} ; 

\textbf{2) Image-Image Alignment:} this approach quantifies similarity between synthetic and real images to filter out low-quality data that deviate from the real distribution \cite{NEURIPS2019_0234c510, lin2023explore, naeem2020reliable, chen2023deep}. Specifically, clustering-based curation~\cite{lin2023explore} use real-data cluster centroids as anchors to retrieve synthetic samples. While straightforward, these methods rely heavily on visual similarity, potentially limiting the diversity of selected data for downstream tasks.
\vspace{3pt}
\newline
\noindent \textbf{Diversity-Guided Curations.} As generative images become increasingly realistic, diversity is key for downstream usage. Prior works enhance diversity by varying prompts~\cite{Shipard_2023}, applying text-conditioned augmentation~\cite{dunlap2023diversify, da2023diversified}, using textual inversion~\cite{trabucco2024effective}, or conditioning on classifier outputs~\cite{hemmat2024feedbackguideddatasynthesisimbalanced}. While effective, these methods require fine-tuning or prompt engineering and focus on generation-time diversity, offering little guidance on how to efficiently leverage existing synthetic datasets.
In this work, we do not intervene in the generation process; instead, we propose a post-generation data curation pipeline that jointly considers fidelity and diversity.

\section{Methods}
\label{sec:Methods}

\noindent\textbf{Overview.}
We target the following \emph{post-generation curation} problem and assume only sample access to the generator. Given (i) a labeled real reference set $\mathcal{D}_R=\{(I_i,y_i)\}_{i=1}^{|\mathcal{D}_R|}$, (ii) a synthetic pool $\mathcal{D}_S=\{\tilde I_j\}_{j=1}^{|\mathcal{D}_S|}$ with class assignments inherited from the generator's conditioning, and (iii) a selection budget $k$ (possibly much larger than $|\mathcal{D}_R|$), the goal is to select $\mathcal{A}\subseteq\mathcal{D}_S$ so that the selected synthetic training set improves downstream classification on the real distribution. 

Our method has two steps. First, we split the real reference images into a \emph{Homogeneous} subset $\mathcal{I}_{HO}$ of local representatives and a \emph{Heterogeneous} subset $\mathcal{I}_{HE}$ of non-redundant variation (Sec.~\ref{Categorizing}).
Second, we allocate the synthetic budget across these two subsets and score each synthetic candidate with a partition-conditioned fidelity--diversity criterion (Sec.~\ref{selection_strategy}). The key design principle is that the selected set should match both the canonical component and the heterogeneous component of the real class, instead of ranking all synthetic images against a single pooled reference distribution.

\subsection{Homogeneous (\ho) and Heterogeneous (\he) Samples}
\label{Categorizing}
To guide the selection of synthetic data, we first categorize target real distribution into two distinct sets: \textbf{Homogeneous (\ho)} and \textbf{Heterogeneous (\he)}. 
\underline{\ho instances} represent the canonical semantics, exhibiting high intra-class similarity in feature space.
\underline{\he instances} capture greater variation, including less typical instances that contribute to diversity.

\smallskip
\noindent\textbf{Identifying \ho and \he Instances.}  
We construct the \ho/\he split from the directed $1$-nearest-neighbor (1-NN) graph of each real class in feature space. The construction is local, non-parametric, and costs $O(n^2d)$ for a class with $n$ examples and $d$ is the feature dimension of input. Unlike a centroid-near/far split (\eg $k$-means, core-set partitions), it does not impose a global axis through the class: $\mathcal{I}_{\mathrm{HO}}$ contains examples that other examples use as their closest local representatives, while $\mathcal{I}_{\mathrm{HE}}$ contains examples with zero in-degree in this local graph.

Given images $\mathcal{I}=\{I_1,\ldots,I_n\}$, we extracted $\ell_2$-normalized features $\mathcal{F}=\{f_1,\ldots,f_n\}$  by MoCo v3~\cite{chen2021mocov3} encoder. We use cosine distance $d(f_i,f_j)=1-\langle f_i,f_j\rangle$ throughout. For each $I_i$, we define its within-class
nearest neighbor
\begin{equation}
j^\ast(i)=\arg\max_{j\neq i}\langle f_i,f_j\rangle .
\label{eq:nn-map}
\end{equation}
Therefore, the \ho set is the image
of this map and the \he set is its complement:
\begin{equation}
\mathcal{I}_{\mathrm{HO}}=\{I_i:\exists j\neq i,\;j^\ast(j)=i\},\qquad
\mathcal{I}_{\mathrm{HE}}=\mathcal{I}\setminus\mathcal{I}_{\mathrm{HO}}.
\label{eq:ho-he}
\end{equation}
Equivalently, in the directed 1-NN graph with edges
$I_i\to I_{j^\ast(i)}$, $\mathcal{I}_{\mathrm{HO}}$ is the set of nodes with positive
in-degree and $\mathcal{I}_{\mathrm{HE}}$ is the set of zero in-degree nodes.

\begin{figure}
  \centering
  \includegraphics[width=0.95\linewidth]{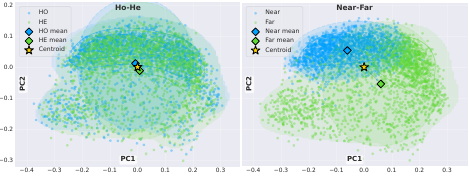}
  \caption{PCA of the CIFAR-10 \textit{horse} under different partitioning rules. The 1-NN \textit{HO-HE split} (right) preserves local neighborhood, whereas \textit{centroid-based split} (left) cuts a class along a single global axis.}
  \label{fig:partition_comp}
\end{figure}

\begin{proposition}[$\mathcal{I}_{HO}$ as a minimal nearest-neighbor cover]
\label{prop:nn-cover-main}
Assume each sample has a unique within-class nearest neighbor.
Then $\mathcal{I}_{HO}$ is the unique inclusion-minimal subset
$A \subseteq \mathcal{I}$ such that:
\begin{equation}
\min_{I_i \in A,\, i \neq j} d(f_j,f_i)
\;=\;
\min_{I_i \in \mathcal{I},\, i \neq j} d(f_j,f_i),
\quad \forall\, I_j \in \mathcal{I}.
\label{eq:nn-cover-main}
\end{equation}
Thus, with
$\varepsilon = \max_j d(f_j,f_{j^\ast(j)})$,
every real sample has an $\mathcal{I}_{HO}$ representative within distance $\varepsilon$.
\end{proposition}

\begin{figure*}[b]
    \centering
    \begin{minipage}{0.395\textwidth}
        \centering
        \includegraphics[width=\linewidth]{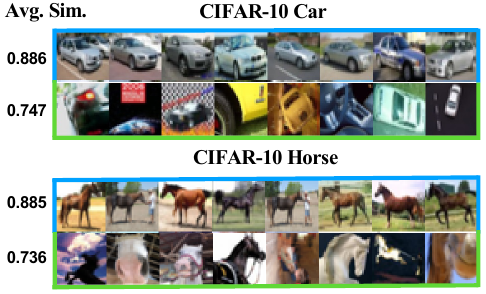} 
        \caption{
        \textcolor{HOColor}{\ho} and \textcolor{HEColor}{\he} instances in CIFAR-10. \textit{Avg. Sim.} is the average similarity between the images in each row and the entire class. \ho data is more representative, expressing the core semantics. \he data is more diverse, capturing a broader range of variations (more cases are in supplementary material).
        }
        \label{fig:ho_he_samples}
    \end{minipage}
    \hfill
    \begin{minipage}{0.58\textwidth}
        \centering
        \includegraphics[width=\linewidth]{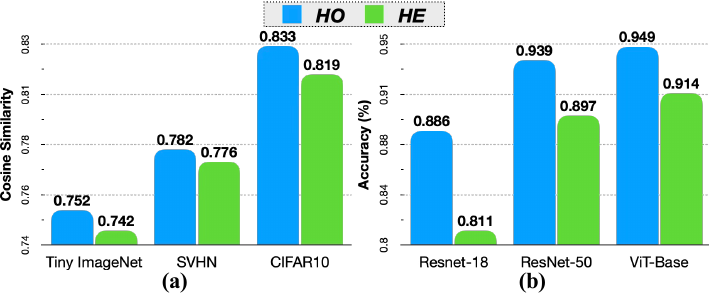}
        \caption{$\boldsymbol{\mathcal{I}_{\mathrm{HO}}} - \boldsymbol{\mathcal{I}_{\mathrm{HE}}}$ downstream impact.   In (a), we measure cosine similarity between \textcolor{HOColor}{\ho}/\textcolor{HEColor}{\he} subsets and generations of models trained by each real dataset. Synthetic data consistently more closely resembles \textcolor{HOColor}{\ho} than \textcolor{HEColor}{\he}. In (b), we evaluate discriminators trained on CIFAR-10. The \textcolor{HOColor}{\ho} subset is consistently better modeled than the \textcolor{HEColor}{\he} subset  across different discriminators. (More empirical results are provided in the supplementary material.)}
        \label{fig:ho_vs_he}
    \end{minipage}
    \label{fig:combined}
\end{figure*}

Proposition~\ref{prop:nn-cover-main} formalizes the representative role of $\mathcal{I}_{\mathrm{HO}}$: replacing the full class by $\mathcal{I}_{\mathrm{HO}}$ preserves the nearest-neighbor reconstruction cost of every training sample. 
The complement $\mathcal{I}_{\mathrm{HE}}$ should therefore not be read as noise or outliers. It is the part of the class that is not needed as a nearest-neighbor representative for other examples, and is precisely where non-redundant variation can be lost when a generator over-produces canonical images.

\noindent\textbf{Empirical $\boldsymbol{\mathcal{I}_{\mathrm{HO}}} - \boldsymbol{\mathcal{I}_{\mathrm{HE}}}$ Analysis and Impact.}\label{preference}
\label{main:graph_theory} 
As visualized in Fig.~\ref{fig:partition_comp}, the \hohe split preserves local neighborhood and keeps each semantic mode mixed across both subsets, avoiding mode imbalance during guided selection. In contrast, centroid-based (clustering) alternatives that define near/far canonical patterns tend to isolate individual mode regions, yielding suboptimal selection references. Without a complete semantic landscape, prioritizing any single mode during synthetic data curation inevitably overlooks parts of the real distribution, leading to information loss. Further visualizing \ho-\he instances in Fig.~\ref{fig:ho_he_samples}. Since \ho samples are retrieved as local representatives, they get higher intra-class similarity than \he samples, both visually and quantitatively.

Analyzing the impact of the $\mathcal{I}_{HO}/\mathcal{I}_{HE}$ partition on downstream performance, although both $\mathcal{I}_{HO}$ and $\mathcal{I}_{HE}$ encompass all  modes (as shown in Fig.~\ref{fig:partition_comp}), we observe an interesting divergence in downstream behaviors: 1) \textbf{GMs preferentially learn and reproduce $\boldsymbol{\mathcal{I}_{HO}}$}. As quantified in Fig.~\ref{fig:ho_vs_he}(a), synthetic data disproportionately mirrors $\mathcal{I}_{HO}$ over $\mathcal{I}_{HE}$. This inherent preference indicates that GMs gravitate toward repeated canonical patterns within $\mathcal{I}_{HO}$, which may explain the persistent diversity limitations observed in modern generative models. 2) \textbf{Discriminators exhibit a similar performance bias, excelling on $\boldsymbol{\mathcal{I}_{HO}}$ while struggling with $\boldsymbol{\mathcal{I}_{HE}}$} shown in Fig.~\ref{fig:ho_vs_he}(b). This discrepancy indicates that, despite sharing underlying semantics, $\mathcal{I}_{HE}$ contains inherently more challenging cases that discriminators fail to model effectively.

Therefore, leveraging synthetic data for downstream training without curation may introduce potential risks. As shown in Proposition~\ref{prop:nn-cover-main}, $\mathcal{I}_{\mathrm{HO}}$ forms a minimal representative subset that preserves the local coverage of the full data distribution. During modeling, generators or discriminators is more likely to prioritize patterns associated with $\mathcal{I}_{\mathrm{HO}}$, as such patterns provide stronger representation of samples than those in $\mathcal{I}_{\mathrm{HE}}$. From an optimization perspective, patterns closer to more samples in the observed distribution generally contribute lower empirical fitting costs. Thus, the optimization process tends to favor modeling $\mathcal{I}_{\mathrm{HO}}$-like patterns. Consequently, synthetic data generation may become biased toward $\mathcal{I}_{\mathrm{HO}}$, and directly using such data for downstream training may further reinforce this inherent distribution bias.

\subsection{Synthetic Data Selection Strategy}
\label{selection_strategy}
Given the \ho and \he sets, we propose a synthetic data selection strategy to identify desirable instances from a fixed synthetic pool. The core objective is to ensure high semantic fidelity while enhancing diversity by prioritizing samples that diverge from the dominant patterns in \ho.

\smallskip

\noindent\textbf{Partitioned Selection Strategy.}
Unlike prior methods that treat the entire real dataset as a single reference for data
selection, we propose a partitioned selection approach. We treat \ho and \he as separate partitions and select synthetic instances by referring to real images within each partition under the same principle: desired synthetics should be sufficiently close to their corresponding real instances, while prioritizing those that deviate from canonical representations. By combining the selection results from \ho and \he instances, we obtain the final selected pool. 

Concretely, we compute a selection score $S^p$ for each synthetic sample based on its correlation with partition $p\in\{\ho, \he\}$. Samples are ranked within each partition, and the instances with high $S^p$ get retrieved. Algorithm~\ref{alg:selection} summarizes the selection procedure.

\begin{algorithm}[!t]

\caption{Data Selection with \ho and \he}
\label{alg:selection}
\begin{algorithmic}[1]
\REQUIRE Feat. maps $\mathcal{F}^{\ho}\!\in\!\mathbb{R}^{n \times d}$, $\mathcal{F}^{\he}\!\in\!\mathbb{R}^{m \times d}$, $\mathcal{F}^{syn}\!\in\!\mathbb{R}^{s \times d}$; centroid $\mathcal{C}^{\ho}\!\in\!\mathbb{R}^{d}$; weight $\alpha \in [0,1]$; size $k$
\ENSURE Selected indices $\mathcal{I}_{\text{selected}}$

\STATE \COMMENT{\textit{Step 1: Construct reference anchors}}
\STATE $\mathcal{R}^{\ho} \gets \mathcal{C}^{\ho}$
\STATE $\mathrm{idx} \gets \operatorname{argmax}(\mathcal{F}^{\he} (\mathcal{F}^{\ho})^\top, \text{dim}=1)$
\STATE $\mathcal{R}^{\he} \gets \mathcal{F}^{\ho}[\mathrm{idx}]$ \COMMENT{Align \he features to \ho}

\STATE \COMMENT{\textit{Step 2: Score calculation and selection}}
\FOR{$p \in \{\ho, \he\}$}
    \STATE $S^p_{\text{fid}} \gets \cos(\mathcal{F}^{syn}, \mathcal{F}^p)$ \COMMENT{Fidelity}
    \STATE $S^p_{\text{div}} \gets -\cos(\mathcal{R}^p - \mathcal{F}^p, \; \mathcal{F}^{syn} - \mathcal{F}^p)$ \COMMENT{Diversity}
    \STATE $S^p \gets  \alpha S^p_{\text{div}} + (1 - \alpha) S^p_{\text{fid}}$
    \STATE $\mathcal{I}^p_{\text{top}} \gets \operatorname{Top}_k(S^p)$
\ENDFOR
\STATE \textbf{return} Selected set $\mathcal{A} \gets \mathcal{I}^{\ho}_{\text{top}} \cup \mathcal{I}^{\he}_{\text{top}}$
\end{algorithmic}
\end{algorithm}

\smallskip
\noindent\textbf{Scoring Mechanism for Each Partition.}
Our selection mechanism balances two competing objectives: the \textit{\underline{fidelity score}}, measuring how closely a synthetic instance resembles the real, and the \textit{\underline{diversity score}}, measuring how much it deviates from canonical patterns in the real dataset.

\textit{1) Fidelity Score.} The fidelity score $S^p_{\text{fid}}$ quantifies how well the synthetic samples align with the real distribution of partition $p$. We measure this using the cosine similarity between the synthetic features $\mathcal{F}^{syn}$ and the real features $\mathcal{F}^p$, as defined in Eq.~\ref{eq:fidelity}. A higher fidelity score indicates that the synthetic samples closely resemble the real data in the corresponding partition:
\begin{equation}
    S^{p}_{\text{fid}} = \cos(\mathcal{F}^{syn}, \mathcal{F}^p).
\label{eq:fidelity}
\end{equation}

\begin{figure}[t] 
    \centering
    \includegraphics[width=0.49\textwidth]{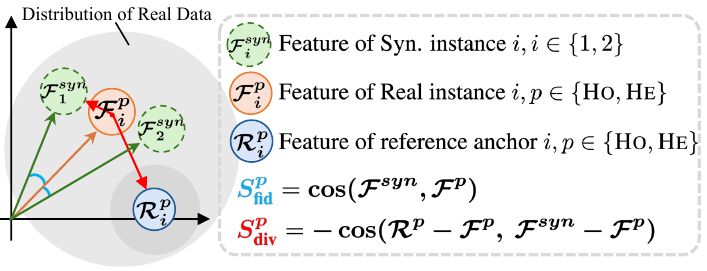}
    \caption{
    Illustration of the computation of $S_\text{fid}^p$ and $S_\text{div}^p$ in partition $p$. Fidelity is assessed as $\cos \left(\mathcal{F}_i^{\text{syn}}, \mathcal{F}^p_{i} \right)$, while diversity is measured by the angle between vectors ending at $\mathcal{F}_i^{\text{syn}}$ and $\mathcal{R}^{p}_{i}$. In the diagram, Syn. instance 1 exhibits greater diversity than Syn. instance 2.
    }
    \label{fig:feature_space_diagram}
\end{figure}

\textit{2) Diversity Score.} The diversity score $S^p_{\text{div}}$ is designed to encourage the selected synthetic samples to deviate from canonical patterns. To achieve this, we define a subset-specific reference anchor $\mathcal{R}^p$. For the \ho subset, the anchor is the centroid of the real features ($\mathcal{R}^{\ho} = \mathcal{C}^{\ho}$). For the \he subset, the anchor is the nearest matching feature in the \ho subset ($\mathcal{R}^{\he} = \mathcal{F}^{\ho}[\mathrm{idx}]$). The diversity metric is then formulated as the negative cosine similarity between two directional vectors, as defined in Eq.~\ref{eq:diversity}: one from the real instance $\mathcal{F}^p$ to the anchor $\mathcal{R}^p$, and the other from the same real instance to the synthetic data $\mathcal{F}^{syn}$.
\begin{equation}
    S^{p}_{\text{div}} = -\cos(\mathcal{R}^p - \mathcal{F}^p, \; \mathcal{F}^{syn} - \mathcal{F}^p).
\label{eq:diversity}
\end{equation}
As illustrated in Fig.~\ref{fig:feature_space_diagram},
if these two vectors point in opposite directions, the cosine similarity is negative, leading to a higher diversity score. If the vectors are aligned, the synthetic sample is closer to the homogeneous distribution, resulting in a lower diversity score.

\textit{3) Total Score.}
To control the trade-off between fidelity and diversity, we define the score as:
\begin{equation}
    {S}^p = \alpha \cdot S_\text{div}^p + (1 - \alpha) \cdot S^p_\text{fid},
\label{f:core_score}
\end{equation}
where $\alpha$ is a hyperparameter that determines the selection priority. $\alpha = 0$ (\textit{MaxSim}) prioritizes fidelity, ensuring that synthetic instances closely resemble real ones. $\alpha = 1$ (\textit{MaxDiv}) prioritizes diversity, encouraging synthetic samples that deviate from the dominant patterns of real data.
Adjusting $\alpha$ enables flexible control over the selection behavior to match downstream needs.

\smallskip
\noindent\textbf{Practical Considerations.}
\textbf{(a) Complexity.} Building the 1-NN graph costs $O(n^2d)$ after feature extraction, where $n$ is number of real samples, $d$ is feature dimension. Scoring is implemented as batched matrix multiplication and costs $O(|\mathcal{D}_S|nd)$, where $|\mathcal{D}_S|$ denotes the number of synthetic candidates.
\textbf{(b) Encoder choice.} The split is relative to encoder $\phi$, so a domain-mismatched encoder can degrade curation. Sec.~\ref{sec:Ana} compares representative encoders and shows that the gains are not tied to a single feature extractor. 
\textbf{(c) No balanced \hohe requirement.} The \hohe split is induced by the 1-NN graph; it does not assume a target ratio or require pre-labeled canonical examples. Unequal partition sizes only change how the synthetic budget is allocated across the two scoring branches: larger partitions receive proportionally more selections, while smaller partitions still receive explicit coverage. Thus, the near-balanced \hohe masses observed in the experiments are an empirical property of the datasets, not a requirement of the method.

\section{Experiments}
\label{sec:Exp_Res}

We evaluate post-generation curation under two settings. The \emph{general} setting fixes a generator and a synthetic pool, applies each selection method under the same budget, and trains downstream classifiers for in-domain (Sec.~\ref{sec:scaling}) and out-of-domain (OOD) (Sec.~\ref{sec:ood}) evaluation. The \emph{plug-in} setting (Sec.~\ref{sec:plugin-post-process-results}) uses the same curation after task-customized generation, testing whether selection remains useful when the generator has already been adapted to a downstream task.

\begin{figure*}[!th]
    \centering
    \includegraphics[width=0.9\linewidth]{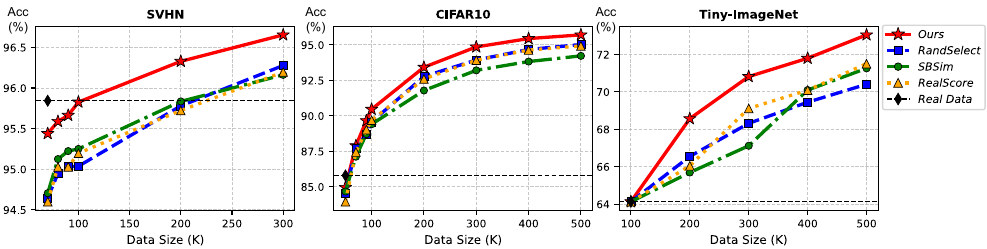}
    \caption{
    Average test accuracy over \textbf{6 runs} for models trained with different amounts of selected synthetic data. We compare with {\textit{RandSelect}}, {\textit{SBSim}}, and {\textit{RealScore}}; \textit{CLIP-Align} is omitted for readability due to poor performance. The full results, including \textit{CLIP-Align}, are provided in supplementary material. The black dashed line denotes the corresponding real-data baseline.
    } 
    \label{fig:scalingup}
\end{figure*}

\subsection{Datasets and Baselines}

\noindent\textbf{Datasets.} 
The main evaluation spans classification and segmentation datasets with increasing scale and visual complexity. For SVHN~\cite{37648}, CIFAR-10~\cite{krizhevsky2010cifar}, and Tiny-ImageNet~\cite{Le2015TinyIV}, covering 10, 10, and 200 classes, respectively, we curate synthetic pools generated by EDM~\cite{wang2023better}. For large-scale validation, we use ImageNet-1K (IN-1K) synthetics generated by EDM2~\cite{Karras2024edm2,Karras2024autoguidance}. Our plug-in experiments further consider task-customized generation on ImageNet-100 (IN-100)\cite{tian2020contrastive} and VOCaug\cite{everingham2015pascal} for classification and segmentation, respectively (Sec.~\ref{sec:plugin-post-process-results}). Finally, to evaluate the robustness of models trained on curated synthetic data, we assess their out-of-distribution generalization across nine OOD benchmarks corresponding to the four in-domain datasets: SVHN, CIFAR-10, Tiny-ImageNet, and IN-1K.
\newline
\noindent\textbf{Baselines.} 
For each comparison, all selection methods operate on the same synthetic pool and use the same selection budget. In the main evaluation, we compare three categories of selection methods. \textit{RandSelect}~\cite{wood2021faketillmakeit,carlini2023extracting,kim2023dcface} is an uninformed random baseline. \textit{CLIP-Align}~\cite{he2023syntheticdatagenerativemodels} represents image--label alignment by ranking generated images with CLIP text--image scores. \textit{RealScore}~\cite{NEURIPS2019_0234c510} and \textit{SBSim}~\cite{lin2023explore} represent image--image alignment.
In the plug-in setting, the baselines are generation-side interventions that already target the downstream task: inference-time intervention~\cite{askari2024improving} and the fine-tuned generator \textit{JoDiffusion}~\cite{wang2026jodiffusion}. In this plug-in setting, we follow the generation configurations specified in the original paper.
\newline
\noindent\textbf{Implementation.} 
Unless otherwise stated, we use $\alpha=0.5$ in Eq.~\ref{f:core_score} for all main experiments and keep the downstream training recipe fixed for fair comparisons. For the feature extration, we use MoCo v3~\cite{chen2021mocov3} as the default extractor. For all experiments, we use NVIDIA L40S GPUs.

\subsection{Scaling Model Performance with Selective Synthetic Data}
\label{Scaling Model Performance with Selective Data}
\label{sec:scaling}

\definecolor{vitblue}{RGB}{220,235,255}
\definecolor{resnetorange}{RGB}{255,235,220}

\begin{table*}[!th]
\centering
\caption{ImageNet-1K Top-1 Acc. (\%) with 1M and 3M selected samples. 
Background colors indicate different models:
\colorbox{vitblue}{ViT-B/16} and \colorbox{resnetorange}{ResNet-50}, 
evaluated under both \textbf{from-scratch} and \textbf{fine-tuning} settings. 
The corresponding training recipes are provided in supplementary material.}
\setlength{\tabcolsep}{4.5pt}
\renewcommand{\arraystretch}{1.1}
\begin{tabular}{l|ccc|ccc|cc|cc}
\hline
\multirow{2}{*}{\textbf{Selection}} 
& \multicolumn{6}{c|}{\textbf{Scratch}} 
& \multicolumn{4}{c}{\textbf{Fine-tune}} \\
\cline{2-9}
 & \cellcolor{vitblue}+0 & \cellcolor{vitblue}+1M & \cellcolor{vitblue}+3M 
 & \cellcolor{resnetorange}+ 0& \cellcolor{resnetorange}+1M & \cellcolor{resnetorange}+3M 
 & \cellcolor{vitblue}1M & \cellcolor{vitblue}3M 
 & \cellcolor{resnetorange}1M & \cellcolor{resnetorange}3M \\
\hline
Random     & 64.22 & 70.90 & 73.54 
           & 69.27 &  72.03     &  72.13
           & 78.52 & 78.68 
           & 71.90 & 73.65 \\
RealScore  & - & 70.92 & 73.53 
           & - & 72.01 & 72.16
           & 78.52 & 78.67 
           & 71.93 & 73.62 \\
CLIP-Align & - &   69.18  &  70.88   
           & - & 69.44 &  70.98
           &  72.40 & 75.23 
           & 62.57     &  67.57  \\
SBSim ($\alpha=0$)& - & 70.95 & 73.23 
           & - & 71.97 &  72.18
           & 78.58 & 78.76 
           & 72.18 & 74.14 \\
\hline
Diversity ($\alpha=1$)  & - & 71.42 & 73.79 
           & - & 71.80 &  73.25
           & 78.72 & 79.14 
           & 72.25 & 74.25 \\
Ours       & - & \textbf{71.49} & \textbf{74.02} 
           & - & \textbf{72.09} &  \textbf{73.33}
           & \textbf{78.97} & \textbf{79.36} 
           & \textbf{73.14} & \textbf{74.76} \\
\hline
\end{tabular}
\label{tab:main_imagenet1k}
\end{table*}

We first ask whether dataset curation improves the value of synthetic utility as the training budget grows. For each selected set, we train the same downstream architecture with the same training recipe and then evaluate on the corresponding real test set. In the main experiments, we employ multiple models across different experimental settings. The small- and medium-scale experiments use ResNet-18/50~\cite{he2016deep} and EfficientNet-B0~\cite{tan2019efficientnet}; For the more complex ImageNet-1K experiments, we further include ViT-B/16~\cite{wu2020visual}.

To simulate and analyze varying levels of dataset complexity in real-world settings, we evaluate our method across \textbf{SVHN}, \textbf{CIFAR-10}, and \textbf{Tiny-ImageNet}. As shown in Fig.~\ref{fig:scalingup}, our curation consistently improves downstream performance, with larger gains as the selection budget increases.
On \textbf{SVHN}, our method matches the real-data baseline with substantially fewer examples (\ie 100\,k synthetic samples), whereas alternative methods require over 200\,k to reach comparable accuracy.  
On \textbf{CIFAR-10}, our method achieves 95\% accuracy with 300\,k samples, while \textit{RandSelect} and \textit{RealScore} require around 500\,k, and \textit{SBSim} demands substantially more. This pattern is consistent with the failure mode targeted by our method: once high-fidelity canonical samples saturate, additional gains depend on selecting less redundant variation.
For the harder \textbf{Tiny-ImageNet}, we use selected synthetic data as augmentation since synthetic-only training suffers a larger domain gap; our curation improves ResNet-50 accuracy at every augmentation budget.
\newline
For \textbf{IN-1K}, prior work~\cite{sariyildiz2023faketillmakeit, azizi2023syntheticdatadiffusionmodels, fan2023scalinglawssyntheticimages} reported a persistent gap between real and synthetic data at ImageNet scale.
 We therefore evaluate two practical uses of selected synthetic data: 1) training models from scratch with synthetic data as augmentation and 2) fine-tuning pre-trained models using synthetic data alone. 
Tab.~\ref{tab:main_imagenet1k} reports the results for ViT-B/16 and ResNet-50 with 1M and 3M selected images from an EDM2 pool. In the training-from-scratch setting, our method consistently achieves superior performance when combined with 1M real images under both +1M and +3M synthetic data settings. Under fine-tuning, using the same training configuration, our method achieves the highest accuracy. Although ImageNet remains challenging due to the persistent real-to-synthetic gap, our results demonstrate the effectiveness of curated synthetic data at scale, showing that the fidelity–diversity balance remains valuable for large-scale visual recognition.

\subsection{Plug-in Curation after Generator-Side Interventions}
\label{sec:plugin-post-process-results}

To maximize the utility of synthetic data, customizing the generators provides a generation-intervention strategy. We further investigate whether post-generation curation remains beneficial after such generator-side improvements, thereby evaluating the complementarity between the two strategies. Keeping each task-adapted pool fixed after customized generations, we apply our selection as a plug-in post-processing step. We study two representative types of generation intervention: inference-time intervention for classification~\cite{askari2024improving} and generator fine-tuning for segmentation~\cite{wang2026jodiffusion}.

\begin{table}[!t]
\centering
\caption{Accuracy and data reduction in ImageNet-100 using ResNet-18.}
\label{tab:data_reduction}
\setlength{\tabcolsep}{5pt}
\renewcommand{\arraystretch}{1.08}
\footnotesize

\begin{tabular}{lccc}
  \toprule
  Method & Data Size & Accuracy (\%) & Reduction \\
  \midrule
  Baseline~\cite{askari2024improving}
    & 50K  & 78.14 & -- \\
  Baseline + Ours
    & \textbf{30K} & \textbf{79.00} & \textbf{40\%} \\
  \midrule
  Baseline~\cite{askari2024improving}
    & 100K & 79.36 & -- \\
  Baseline + Ours
    & \textbf{50K} & \textbf{79.48} & \textbf{50\%} \\
  \bottomrule
\end{tabular}
\end{table}

\begin{table}[!t]
    \centering
    \caption{Segmentation performance with and without post-generation
    curation at different synthetic-data scales. Results are reported in terms of mIoU (\%).}
    \label{tab:data_size_comparison_in_segmentation}
    \setlength{\tabcolsep}{7pt}
    \renewcommand{\arraystretch}{1.08}
    \footnotesize

    \begin{tabular}{lcc}
      \toprule
      Method & 10K Images & 15K Images \\
      \midrule
      JoDiffusion~\cite{wang2026jodiffusion}
        & 73.61 & 73.87 \\
      JoDiffusion + Ours
        & \textbf{74.94} & \textbf{76.04} \\
      \bottomrule
    \end{tabular}
\end{table}

\noindent\textbf{Classification.}
We apply our curation to synthetic images generated with inference-time intervention~\cite{askari2024improving} in IN-100, then train ResNet-18 on the selected data. As shown in Tab.~\ref{tab:data_reduction}, our curation further improves synthetic data utility, yielding slight performance gains while reducing task-aware samples by 40–50\%.

\noindent\textbf{Segmentation.}
We curate synthetic VOCaug data produced by the fine-tuned generator JoDiffusion~\cite{wang2026jodiffusion}, then train the downstream segmentation model (\ie DeepLabV3~\cite{chen2018encoder
}). In Tab.~\ref{tab:data_size_comparison_in_segmentation}, we report mIoU under the same training data scale, showing that our method provides additional gains beyond customized generation via fine-tuning. 

Overall, results across both classification and segmentation show that post-generation curation is a generator-agnostic approach that further improves the utility of optimized synthetic data. Notably, the gains become more pronounced as training volume increases (Tabs.~\ref{tab:data_reduction} and \ref{tab:data_size_comparison_in_segmentation}). We attribute this to, without curation, increasing redundancy and mode collapse at larger synthetic data scales, which limit the benefits of generation-side optimization alone. Post-generation curation therefore provides an increasingly effective complement to customized generation as synthetic data scale grows.

\subsection{Enhancing Model Generalizability with Selected Samples}
\label{sec:ood}

Beyond the in-domain evaluation of both general and customized generation, we further investigate whether our curation strategy enhances model robustness under distribution shifts. As informative and diverse training data are expected to promote better generalization, we evaluate the best models trained from scratch in Sec.~\ref{sec:scaling} on their corresponding OOD benchmarks. The results demonstrate that training on curated data consistently improves out-of-distribution generalization.

\begin{figure}[!t]
    \centering
    \includegraphics[width=1.01\linewidth]{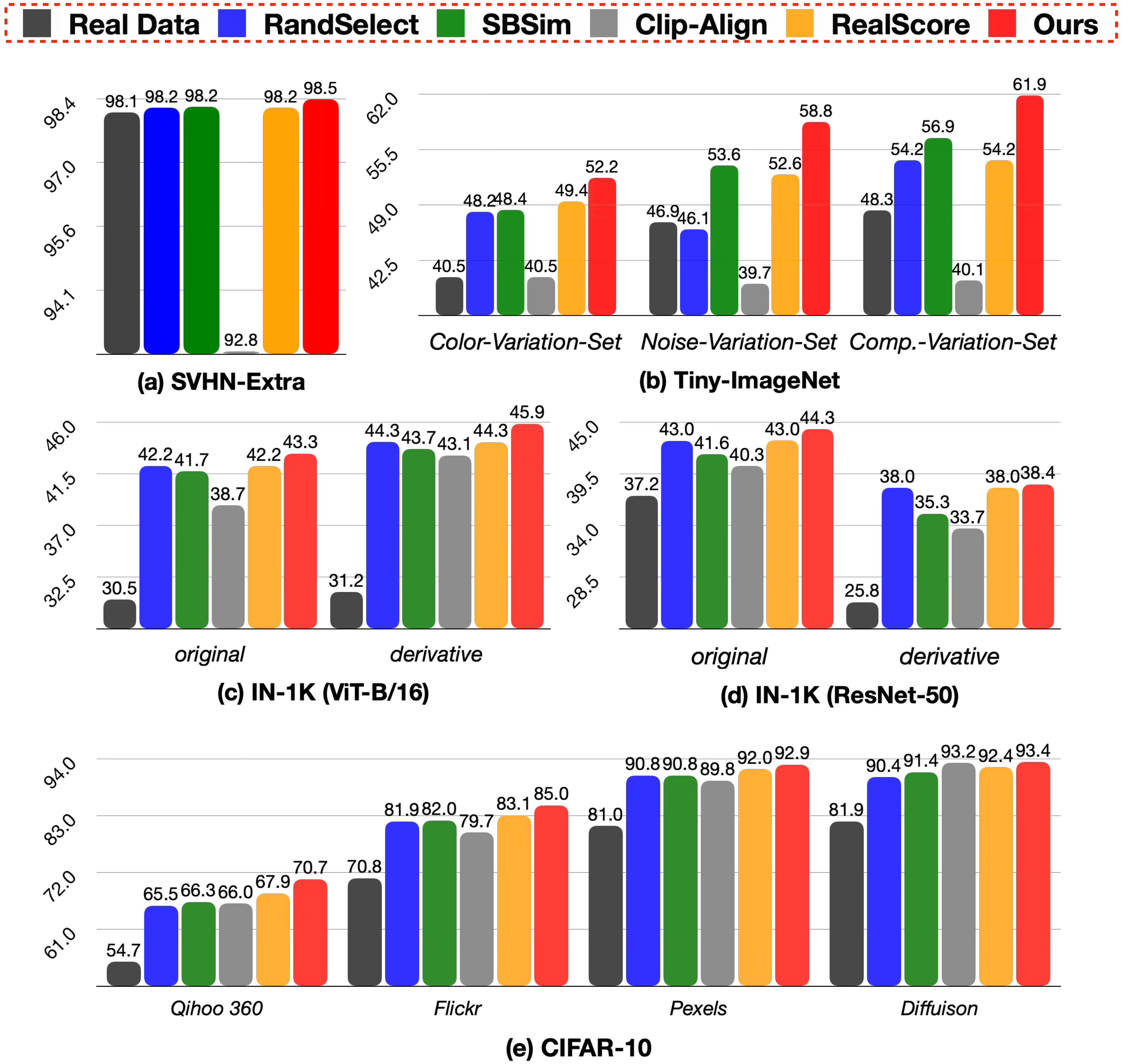}
    \caption{OOD Evaluation. 
    Among different settings, from simple SVHN to chanllenging ImageNet, models trained by our curated data pools consistently achieves the best performance in OOD settings.}
    \label{fig:ood_results}
\end{figure}

\noindent\textbf{OOD of SVHN.} We complement our evaluation with OOD testing in the SVHN extra split~\cite{netzer2011reading}, with the results shown in Fig.~\ref{fig:ood_results}(a). 
\noindent\textbf{OOD of Tiny-ImageNet.} 
We use Tiny-ImageNet-C~\cite{hendrycks2018benchmarking}, which incorporates various types of corruption. We classify them into three types: color-variation set (\ie brightness adjustment, contrast variation), noise-variation set (\ie pixelation, Gaussian noise, motion blur), and compression-variation set (\ie JPEG compression). The results are reported in Fig.~\ref{fig:ood_results}(b).
\noindent\textbf{OOD of IN-1K.} We use several OOD datasets and categorize them into two groups 1) \textit{original suite}, including ImageNet-V2~\cite{recht2019imagenetclassifiersgeneralizeimagenet} and ImageNet-Sketch~\cite{wang2019learning}; 2) \textit{derivative suite}, including ImageNet-C~\cite{hendrycks2019robustness}, -Drawing, and -Cartoon~\cite{salvador2022imagenetcartoon}, derived from the validation set. The evaluation results are summarized in Fig. \ref{fig:ood_results}(c), (d).
\noindent\textbf{OOD of CIFAR-10.} 
CIFAR-10-Warehouse~\cite{Cifar-10-warehouse} is used as a benchmark, collecting data from diverse sources (\eg different search engines). The results are reported in Fig.~\ref{fig:ood_results}(e).

\section{Analysis and Discussion}
\label{sec:Ana}

In this section, we provide further analyzes and discussions about post-generation curation to better understand the proposed curation pipeline, validate its design, and demonstrate the practical significance.

\subsection{Characteristics of the \hohe Split in Real Data}
\label{A}

Our curation strategy is built upon the partition of real data into $\mathcal{I_{HO}}$ and $\mathcal{I_{HE}}$; therefore, understanding their respective characteristics is essential. In Sec.~\ref{sec:Exp_Res}, 1). we observe that the $\mathcal{I}_{HO}$ and $\mathcal{I}_{HE}$ subsets remain nearly balanced across classes and datasets, suggesting that {approximately half of the samples effectively capture the primary semantics of each class.}
2). Using $\mathcal{I_{HO}}$ and $\mathcal{I_{HE}}$ as curation references, we observed that although the two sets have comparable volumes, $\mathcal{I}_{HO}$ retrieves fewer unique synthetic instances, indicating that multiple distinct $\mathcal{I}_{HO}$ samples tend to highly score the same synthetic samples.
We attribute this to the {higher intra-class similarity within $\mathcal{I}_{HO}$, as shown in Tab.~\ref{tab:ho_he_intr_class}, which makes it more challenging for generators to capture and reproduce fine-grained variations based on such similar samples than the more diverse real data in $\mathcal{I}_{HE}$}.

\begin{table}[!t]
\centering
\caption{Average intra-class similarity of $\mathcal{I}_{HO}$ and $\mathcal{I}_{HE}$ across datasets. $\mathcal{I}_{HO}$ consistently gets higher intra-class similarity.}
\label{tab:intra_class_similarity}
\begin{tabular}{lcccc}
\toprule
Subset & SVHN & CIFAR-10 & Tiny-ImageNet & ImageNet-1K \\
\midrule
$\mathcal{I}_{HO}$ & 0.8507 & 0.7948 & 0.8236 & 0.8000 \\
$\mathcal{I}_{HE}$ & 0.8385 & 0.7887 & 0.7894 & 0.7725 \\
\bottomrule
\end{tabular}
\label{tab:ho_he_intr_class}
\end{table}

\subsection{CLIP Filtering Biases Toward Canonical Modes}
Motivated by the significantly poorer performance of \textit{CLIP-Align} compared to other counterparts in Secs.~\ref{sec:scaling} and \ref{sec:ood}, we further investigate its underlying behavior. Despite its widespread use, the limited performance of \textit{CLIP-Align} suggests that such alignment-based filtering may be insufficient or even ineffective for curating modern, high-quality synthetic data, due to its strong bias toward canonical modes.
In Tab.~\ref{tab:feature_similarity}, we provide a quantitative diagnostic on CIFAR-10: across CLIP, MoCo v3, and ConvNeXt feature encoders, high-CLIP-score samples selected by \textit{CLIP-Align} are consistently closer to $\mathcal{I}_{HO}$ than to $\mathcal{I}_{HE}$. Therefore, the \textit{CLIP-Align} data pool is biased toward canonical modes, failing to capture the full distribution represented by both $\mathcal{I}_{HO}$ and $\mathcal{I}_{HE}$ equally. This bias can lead to overfitting to canonical modes, degrading both in- and out-of-domain performance. This issue stems from the preference of CLIP and other discriminator-based selection methods for canonical modes in the training phase optimization. In contrast, our curation strategy relies on the full data distribution comprising both $\mathcal{I}{HO}$ and $\mathcal{I}{HE}$, treating samples from both subsets equally. The hyperparameter $\alpha$ controls the proximity of selected synthetic samples to each real sample, thereby preserving the distribution of $\mathcal{I}_{HE}$ and mitigating the potential underrepresentation that may arise from off-the-shelf filtering methods.

\begin{table}[!t]
    \centering
    \caption{Mean feature similarity between high-CLIP-score synthetic samples and the $\mathcal{I}_{\mathrm{HO}}$ and $\mathcal{I}_{\mathrm{HE}}$ subsets.}
    \label{tab:feature_similarity}
    \setlength{\tabcolsep}{5pt}
    \renewcommand{\arraystretch}{1.08}
    \footnotesize
    \begin{tabular}{lccc}
        \toprule
        & \multicolumn{3}{c}{Feature Encoder} \\
        \cmidrule(lr){2-4}
        Reference Subset & CLIP & MoCo & ConvNeXt \\
        \midrule
        $\mathcal{I}_{\mathrm{HO}}$ & 0.946 & 0.934 & 0.839 \\
        $\mathcal{I}_{\mathrm{HE}}$ & 0.943 & 0.914 & 0.805 \\
        \bottomrule
    \end{tabular}
\end{table}

\begin{figure}[!t]
    \centering
    \includegraphics[width=\linewidth]{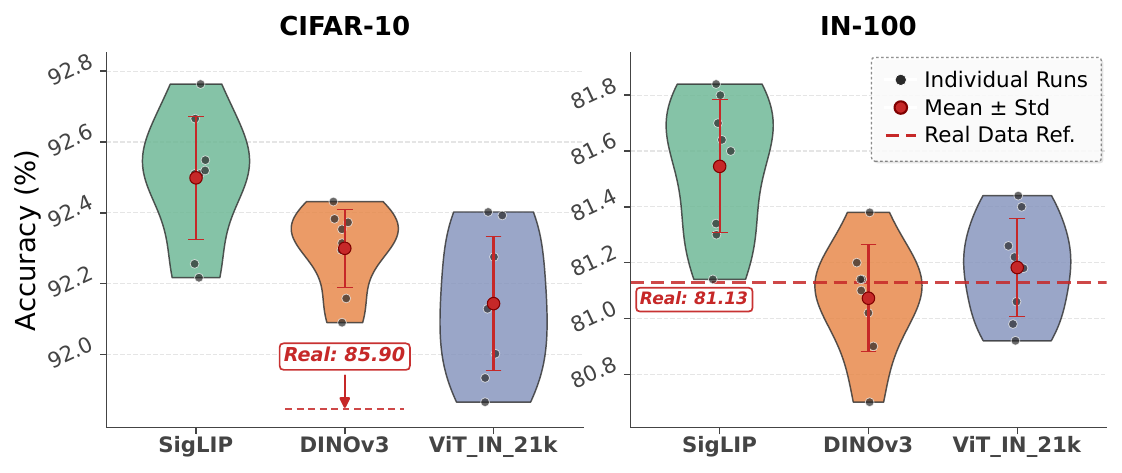}
    \caption{Violin plots of test accuracy after training on selected synthetic data using different feature extractors. Despite variations in mean accuracy, the performance distributions largely overlap on both settings, indicating the robustness of the proposed curation strategy across feature extractors. Detailed training configurations are provided in the supplementary material.}
    \label{fig:different_encoders}
\end{figure}

\subsection{Limited Impact of Feature Extractor Choice}
Since our curation relies on extracted image features, we examine whether the choice of feature extractor matters. We conduct an ablation study using three representative encoders, each reflecting a distinct training paradigm: SigLIP (text–image alignment)~\cite{zhai2023sigmoid}, DINOv3 (self-supervised)~\cite{simoni2025dinov3}, and ViT (supervised). Evaluations are performed on two datasets with different image resolutions and complexity, CIFAR-10 and ImageNet-100. After feature extraction and construction of the curated datasets, we repeatedly train the downstream model with different random initializations, with the results summarized in Fig.~\ref{fig:different_encoders}. In the results, SigLIP consistently achieves the best performance, suggesting that it is a promising choice for feature extraction. However, the substantial overlap in performance distributions across repeated trials, as shown in Fig.~\ref{fig:different_encoders}, indicates that the choice of feature extractor has a relatively minor impact compared with synthetic data quality and $\alpha$ selection in the proposed strategy.

\begin{figure*}[!t]
    \centering

    \includegraphics[width=0.8\textwidth]
    {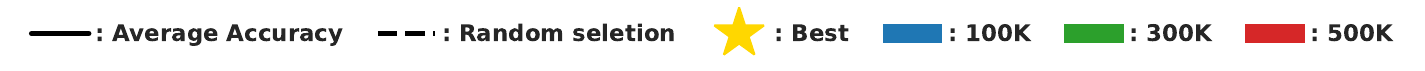}
    \par\vspace{-0.4em}
    \subfloat[StyleGAN~\cite{Karras2020ada} (FID: 3.17)
      \label{fig:tradeoff-stylegan}]{%
      \includegraphics[width=0.325\textwidth]
      {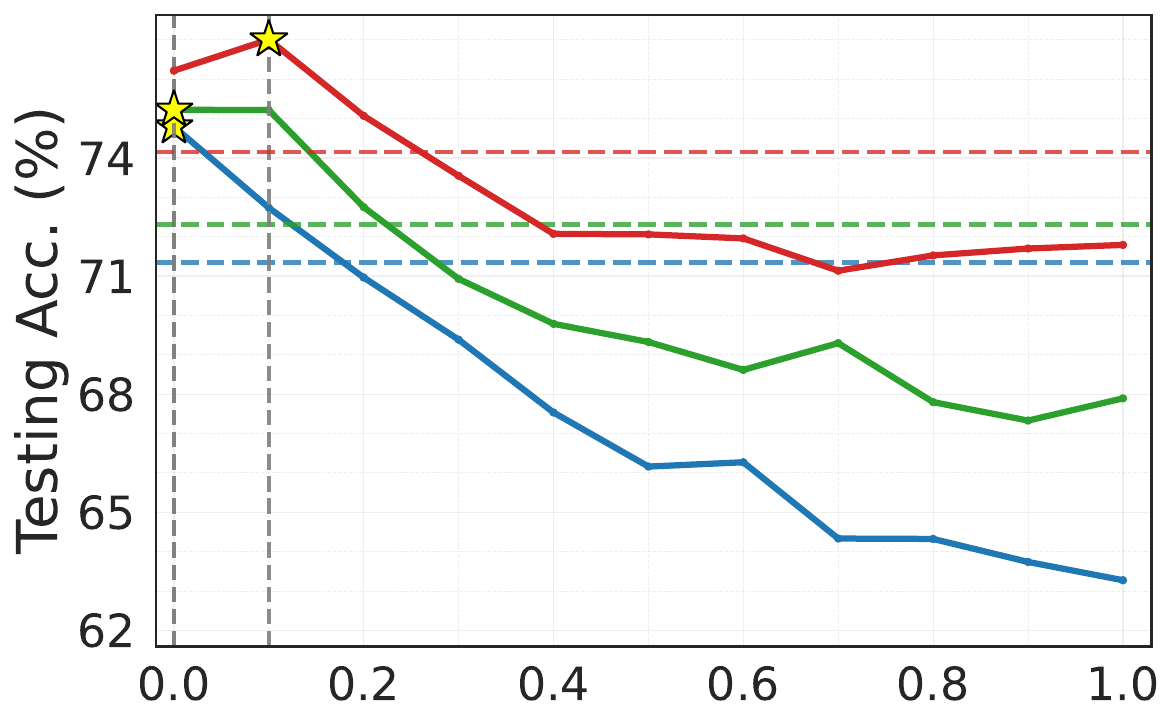}
    }%
    \hfill
    \subfloat[EDM~\cite{Karras2022edm} (FID: 2.67)
      \label{fig:tradeoff-edm}]{%
      \includegraphics[width=0.325\textwidth]
      {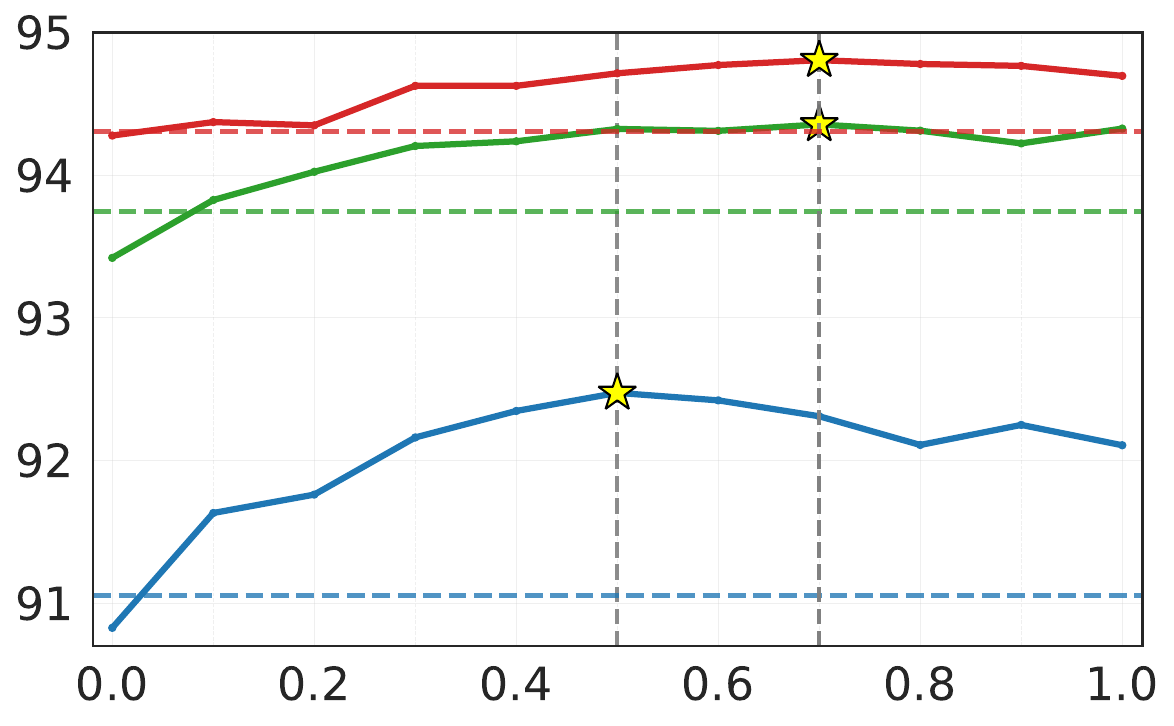}
    }%
    \hfill
    \subfloat[EDM2~\cite{Karras2024edm2} (FID: 1.39)
      \label{fig:tradeoff-edm2}]{%
      \includegraphics[width=0.325\textwidth]
      {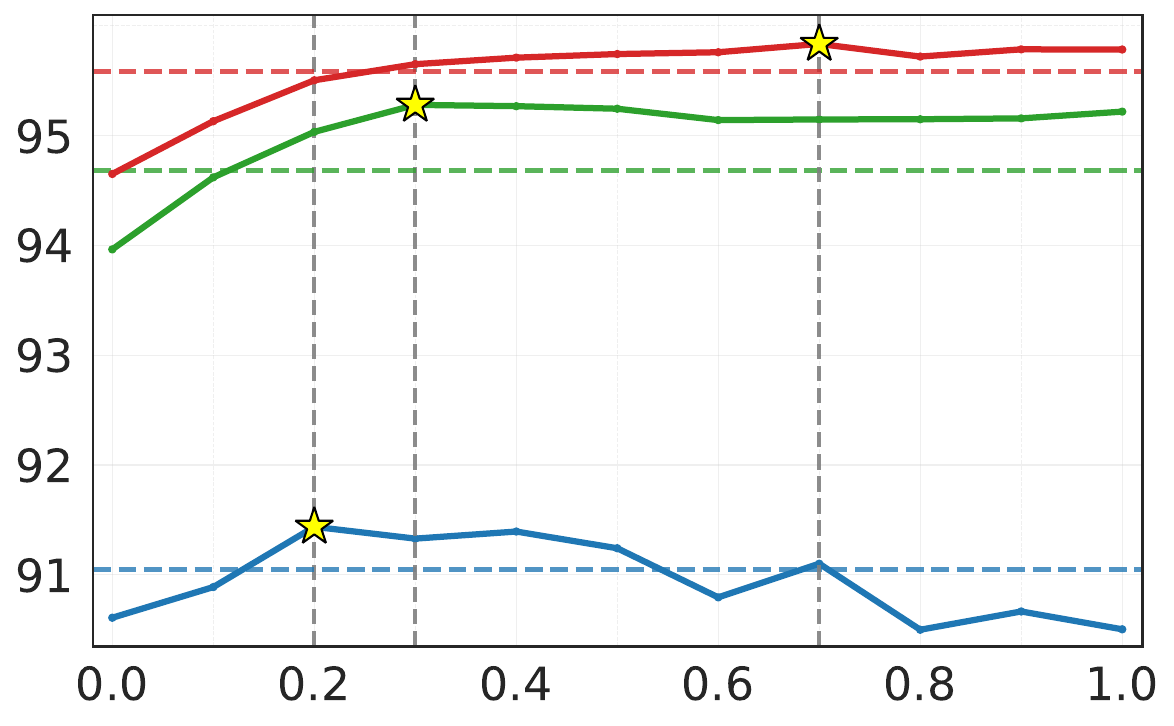}
    }

    \caption{Classification accuracy versus the trade-off parameter $\alpha$,
    where $\alpha=0$ emphasizes fidelity and $\alpha=1$ emphasizes diversity.
    Each subfigure compares the 100K, 300K, and 500K settings with fitted
    trend lines; the optimal point on each curve is marked by a star.
    Results are obtained using ResNet-18 and reported as the average accuracy
    over eight runs.}
    \label{fig:tradeoff-all}
  \end{figure*}

\subsection{Generator Quality and Data Budget Shift the Optimal Fidelity-Diversity Balance ($\boldsymbol{\alpha}$).}
Our curation strategy employs the hyperparameter $\alpha$ in Eq.~\ref{f:core_score}. Here, we investigate how $\alpha$ interacts with different generators and affects downstream discriminative performance. Our experiments use 1M synthetic CIFAR-10 samples from three generators with varying generation quality. As shown in Fig.~\ref{fig:tradeoff-all}, the optimal $\alpha$ depends on both the quality of the synthetic data and the volume of training data. (1) For StyleGAN2 in Fig.~\ref{fig:tradeoff-stylegan}, which yields the highest FID (poorest generation quality), fidelity remains the key factor for classification performance, with the optimal $\alpha$ close to 0 among different scale settings; (2) When the generated distributions are closer to the real in EDM and EDM2, Figs.~\ref{fig:tradeoff-edm}, \ref{fig:tradeoff-edm2} demonstrate the benefit of tuning the trade-off factor $\alpha$; (3) Across all three generators, as the training volume increases, the optimal $\alpha$ shifts toward larger values, underscoring the increasing importance of diversity at larger scales, where canonical high-fidelity samples provide limited additional gains. 
(4) Regarding the optimal $\alpha$ for each generator, the largest shift is observed for the more advanced EDM2, suggesting that, with larger training volumes and highly realistic generations, diversity becomes increasingly important.

\subsection{Method-Specific Curation Overhead Is Small}
\label{small_curation_overhead}

\begin{table}[!t]
    \centering
    \caption{Runtime overhead of curating a 10M-image synthetic pool at different real-data scales.}
    \label{tab:runtime_scale}
    \setlength{\tabcolsep}{4pt}
    \renewcommand{\arraystretch}{1.08}
    \footnotesize
    \begin{tabular}{lccc}
        \toprule
        Stage & 10K & 100K & 1M \\
        \midrule
        Feature extraction
            & 65.07 Min. & 65.71 Min. & 75.55 Min. \\
        \hohe splitting
            & 0.51 Sec. & 1.06 Sec. & 18.42 Sec. \\
        Scoring and selection
            & 5.41 Sec. & 80.10 Sec. & 821.81 Sec. \\
        \bottomrule
    \end{tabular}
\end{table}

Method-specific curation overhead is an important factor in assessing the practical utility of our curation pipeline. Compared to previous post-generation selection, our curation introduces unique cost in $\mathcal{I}_{HO}/\mathcal{I}_{HE} $ splitting, scoring and selection. In practice, once features are extracted, all such operations reduce to batched cosine similarity computations that are GPU-friendly and fast in practice (implementation in supplementary material). Tab.~\ref{tab:runtime_scale} shows the runtime of each stage of our curation for 10M synthetic samples across varying real-data scales. During curation, comparison among three stages, feature extraction dominates the overall overhead (from 65.07 to 75.55 minutes). Since it is a one-time inference cost shared by most instance-selection baselines, the overhead specific to our method about $\mathcal{I}_{HO}/\mathcal{I}_{HE} $ splitting, scoring and selection remain negligible even at large scales.




\subsection{Practical Significance in Downstream Training}

\begin{table}[!t]
    \centering
    \caption{Overall training overhead comparison on Tiny-ImageNet using
    EfficientNet-B0 and one NVIDIA L40S GPU. Runtime is reported as
    downstream training $+$ curation overhead.}
    \label{tab:compute_compare}
    \setlength{\tabcolsep}{4pt}
    \renewcommand{\arraystretch}{1.08}
    \footnotesize
    \begin{tabular}{lcccc}
        \toprule
        Method & Data Size & \multicolumn{2}{c}{Runtime (Min.) $\downarrow$} & Accuracy (\%) $\uparrow$ \\
        \cmidrule(lr){3-4}
        & & Train $+$ Curation & Total & \\
        \midrule
        Random & 300K & $120 + 0$   & 120 & 74.59 \\
        Ours   & 200K & $80 + 3.8$  & \textbf{83.8}  & \textbf{75.03} \\
        \midrule
        Random & 500K & $210 + 0$   & 210 & 75.70 \\
        Ours   & 300K & $120 + 3.8$ & \textbf{123.8} & \textbf{75.91} \\
        \bottomrule
    \end{tabular}
\end{table}

The practical significance of our data curation is reflected not only in the final model performance but also in the overall training efficiency.As discussed in Sec.~\ref{small_curation_overhead}, our method introduces only limited curation-specific overhead. When considering the entire pipeline from data curation to model training, the reduced training volume more than compensates for this additional cost, leading to lower overall computational cost. 

Taking Tiny-ImageNet setting used in Secs.\ref{sec:scaling} and\ref{sec:ood} as an example, curating a 1M-image synthetic pool takes only 3.8 minutes, including 3.5 minutes for feature extraction and 0.3 minutes for scoring and selection. In comparison, training on 200K, 300K, and 500K synthetic samples takes 80, 120, and 210 minutes, respectively, on a single NVIDIA L40S GPU. As shown in Tab.~\ref{tab:compute_compare}, even after accounting for the additional 3.8-minute curation cost, our method reduces the total runtime by 30.2\% and 41.0\% compared with random selection, respectively, while improving classification accuracy. These results further highlight the practical importance of data curation before downstream training. Since the curation cost remains fixed while training cost increases with model scale and the number of training epochs, the computational savings of our curation strategy become more pronounced as downstream training becomes more expensive.



\section{Conclusions}
\label{sec:Cons}
We introduced a generator-agnostic framework for post-generation curation of synthetic images. By splitting real data into \ho and \he references and
selecting synthetics with a partition-wise fidelity--diversity score, the method improves in-domain accuracy, strengthens OOD generalization, reduces the amount of synthetic data needed to match real-data baselines, and remains complementary to task-customized generation for both classification and segmentation. The analysis further suggests a practical principle for synthetic-data use: as generator fidelity and training budgets increase, curation should shift from merely retaining recognizable samples toward allocating more budget to non-redundant variation. 


\section{Limitations}
\label{limitation}
This work has several limitations, which also point to promising directions for future research.

\noindent\textbf{1) Dependence on generator quality.}
Our method operates as a post-generation curation strategy built on top of off-the-shelf generators, without intervening in the generation process. As a result, the upper bound of synthetic data utility is inherently constrained by the quality of the generator. When generators are too weak to produce realistic images, best curation may reduce to selecting higher-fidelity samples, while diversity-aware selection become less beneficial due to the negative impact of low-fidelity samples on downstream performance. Nevertheless, improvements in modern generative models make synthetic images increasingly realistic, further enhancing the value of post-generation curation for downstream applications.

\noindent\textbf{2) Limited to unimodal image settings.}
Our current method focuses on curating synthetic image data and does not extend to multimodal settings. As vision-language models continue to advance, synthetic datasets that support multimodal training are becoming increasingly important. However, multimodal curation involves data formats beyond images and requires modeling interactions across modalities, which are not addressed by the current framework. Extending post-generation curation to multimodal data is therefore an important and meaningful direction for future work.

\noindent\textbf{3) Reliance on reference data.}
Our method requires a reference from the target distribution to construct the $\mathcal{I}_{HO}$/$\mathcal{I}_{HE}$ set, following common reference-based curation practices. In practice, such targeted reference is often available, and only a small reference set is sufficient, since the partition relies more on relative intra-class similarity than on dataset size.

\noindent\textbf{4) Dependence on pretrained feature extractors.}
Our scoring procedure relies on pretrained feature extractors. Although Fig.~9 shows that the method is only modestly sensitive to the choice of encoder, a severely domain-mismatched feature extractor may still degrade partition quality. In practice, we recommend using strong general-purpose encoders, such as SigLIP, when domain-specific alternatives are unavailable.

\bibliographystyle{IEEEtran}
\bibliography{Secs/reference}

\end{document}